\documentclass[conference]{IEEEtran}
\IEEEoverridecommandlockouts
\usepackage{cite}
\usepackage{amsmath,amssymb,amsfonts}
\usepackage{algorithmic}
\usepackage{graphicx}
\usepackage{textcomp}
\usepackage{xcolor}
\usepackage{fancyhdr}

\usepackage[utf8]{inputenc}		
\usepackage{color}				
\usepackage{microtype} 			
\usepackage{array}

\def\BibTeX{{\rm B\kern-.05em{\sc i\kern-.025em b}\kern-.08em
    T\kern-.1667em\lower.7ex\hbox{E}\kern-.125emX}}

\begin{document}

\title{A writer-independent approach for offline signature verification using deep convolutional neural networks features\\
\thanks{\textcolor{red}{ This article has been accepted for publication in BRACIS 2018 but has not yet been fully edited. Some content may change prior to final publication.}}
}

\author{\IEEEauthorblockN{Victor L. F. Souza}
\IEEEauthorblockA{\textit{Centro de Informática} \\
\textit{Universidade Federal de Pernambuco}\\
Recife, PE, Brazil \\
vlfs@cin.ufpe.br}
\and
\IEEEauthorblockN{Adriano L. I. Oliveira}
\IEEEauthorblockA{\textit{Centro de Informática} \\
\textit{Universidade Federal de Pernambuco}\\
Recife, PE, Brazil \\
alio@cin.ufpe.br}
\and
\IEEEauthorblockN{Robert Sabourin}
\IEEEauthorblockA{\textit{LIVIA, École de Technologie Supérieure} \\
\textit{Université du Québec}\\
Montreal, Canada \\
robert.sabourin@etsmtl.ca}


}

\maketitle

\begin{abstract}
The use of features extracted using a deep convolutional neural network (CNN) combined with a writer-dependent (WD) SVM classifier resulted in significant improvement in performance of handwritten signature verification (HSV) when compared to the previous state-of-the-art methods. In this work it is investigated whether the use of these CNN features provide good results in a writer-independent (WI) HSV context, based on the dichotomy transformation combined with the use of an SVM writer-independent classifier. The experiments performed in the Brazilian and GPDS datasets show that (i) the proposed approach outperformed other WI-HSV methods from the literature, (ii) in the global threshold scenario, the proposed approach was able to outperform the writer-dependent method with CNN features in  the Brazilian dataset, (iii) in an user threshold scenario, the results are similar to those obtained by the writer-dependent method with CNN features.
\end{abstract}


\section{Introduction}

Signature Verification (SV) systems are used to automatically recognize whether the signature provided by an user actually belongs to the individual who he/she claims to be. Therefore, these systems are useful in many real-world applications, such as credit card transactions or document authentication. Specifically, the problem of automatic Handwritten Signature Verification (HSV) can be defined as follows: given a learning set containing genuine signatures of a set of users, a model is trained to classify the signatures as genuine or forgeries. Genuine signatures are those that really belong to the indicated user; in turn, forgeries are those created by other people \cite{hafemann_review:17}.

The signatures to be verified by the HSV systems can be acquired in two ways: offline (static) and online (dynamic). In offline SV, the signature is acquired after the writing process is completed. In this case, the signature is treated as an image. On the other hand, in online verification, a device is used to collect data as it is produced. So, online models can obtain additional information from the users during writing to perform the verification, such as the position or slope of the pen, or the writing pressure \cite{hafemann_review:17}.

Another important point that deserves to be highlighted when dealing with the HSV problem is the model's user horizon. If a model is trained for each user, the system is called writer-dependent (WD). In this case, initially, a training set is constructed as follows: genuine signatures of the tested user are treated as positive instances and signatures from other users as negative. Next, a binary classifier is trained for each user. Although WD systems achieve good results for the HSV task, requiring a classifier for each user increases the complexity and the cost of the system operations as more users are added \cite{kumar:16}.

On the other hand, HSV systems used to classify signatures of any available user in the dataset are known as writer-independent (WI) systems. In this context, a single model is trained for all users from a dissimilarity space. Thus, the classification inputs are vectors of dissimilarity, which represent the difference between the features of a queried signature and a reference signature of the user. When compared to the WD approach, WI systems are less complex, but in general obtain worse results \cite{hafemann:17}.

Recently, Hafemann et al. \cite{hafemann:17} proposed an approach to deal with the offline HSV problem that uses concepts from both WI and WD systems. The approach carries out feature learning from the signature images in a WI format, using a deep convolutional neural network (CNN) called SigNet. After being trained, the CNN is used to extract representative features from the signatures, which are used to train a writer-dependent Support Vector Machine (SVM) classifier for each writer. Their results showed a significant improvement in performance when compared to the previous state-of-the-art methods.

The main objective of this paper is to investigate whether the deep CNN features learned by the Hafemann et al. model \cite{hafemann:17} (available online\footnote{http://en.etsmtl.ca/Unites-de-recherche/LIVIA/Recherche-et-innovation/Projets/Signature-Verification}) can also lead to good results in a writer-independent HSV context. To this end, it is proposed the use of dichotomy transformation \cite{rivard:13} combined with an SVM as a writer-independent classifier to perform the signature verification. 

The following points will be analyzed: (i) which partial decisions fusion rule is the best (functions max, mean, median and min are tested). (ii) The influence of the number of signatures used in the reference set. (iii) A comparison  with other studies from the literature. The experiments are carried out using the GPDS and the Brazilian PUC-PR datasets.

The remaining of this paper is organized as follows: Section \ref{sec:related_works} presents the related work on signature verification, separating WD and WI approaches. Section \ref{sec:proposed_method} details the proposed method, and Section \ref{sec:experiments} describes the used experimental protocol and the discussion of the results. Lastly, Section \ref{sec:conclusion} concludes the paper and discusses future works.

\section{Related works}
\label{sec:related_works}

In general, two approaches are used for offline handwritten signature verification, writer-dependent (WD) and writer-independent (WI). In the WD scenario, a classifier is trained for each writer and is responsible for authenticating his/her signatures. In the WI context a single classifier is trained for all writers and is responsible for associating the questioned entry signatures to one or more reference signatures in a dissimilarity space \cite{hafemann:17}. Most HSV systems presented in literature follow the WD approach \cite{hafemann_review:17}.

Batista et al. \cite{batista:12} proposed a hybrid generative–discriminative ensemble of classifiers which dynamically selects the classifiers for building a writer-dependent HSV system. During the generative stage, the signatures are divided in a grid format and multiple discrete left-to-right Hidden Markov Models (HMMs) are trained with different number of states and codebook sizes, to be able of working at different levels of perception. Then the HMM likelihoods for each enrolled signature are computed and grouped into a feature vector that is used through a specialized Random Subspace Method to build a pool of two-class classifiers (discriminative stage). For the verification task, the authors propose a new dynamic selection strategy based on the K-nearest-oracles (KNORA) algorithm and on Output Profiles to select the most accurate ensemble to classify the given signature \cite{batista:12}.

Guerbai et al. \cite{guerbai:15} proposed a writer-dependent HSV system based on One-Class SVM (OC-SVM) that tries to reduce the difficulties of having a large numbers of users. As a one-class classification problem, the proposed approach models only one class (genuine signatures). Which is a good characteristic, as, in general, the system only has the genuine signatures for each writer to train the classifier. Nevertheless, the low number of genuine signatures is still an important challenge \cite{guerbai:15}.

In the WI scenario, Bertolini et al. \cite{bertolini:10} proposed a  writer-independent approach for handwritten signature verification. This approach applies the ideas of dissimilarity representation and SVMs as classifiers \cite{bertolini:10}. The two main contributions of the authors are the following: (i) introduce a new graphometric feature set based on the curvature of the most important segments, simulated by using Bezier curves. (ii) The use of an ensemble of classifiers structure to improve the resistance against forgeries. This ensemble is built using a standard genetic algorithm and a pool of base classifiers trained with four different graphometric feature sets \cite{bertolini:10}.

Also, Rivard et al. \cite{rivard:13} have proposed a  writer-independent approach that combines multiple feature extraction, dichotomy transformation, and boosting feature selection. The authors report that the accuracy and reliability of the system can be improved by integrating features from different sources of information, so initially they employ some techniques to extract features at different scales. Then they use the Dichotomy Transformation, which reduces the pattern recognition problem to a 2-class problem. 
A good point that deserves to be highlighted is that with this transformation the system alleviates the challenges of deal with limited number of reference signatures from a large number of users. Finally, an ensemble is built using boosting feature selection that uses low-cost classifiers capable of automatically select relevant features during training \cite{rivard:13}. 

Some authors use a combination of both WD and WI approaches. For
example, Eskander et al. \cite{eskander:13} proposed a hybrid writer-independent-writer-dependent model. The aim of the authors was to maximize the positives of each approach. In the scenario where only a few genuine signatures are available, they use the writer-independent classifier to perform the verification. On the other hand, the writer-dependent classifier is trained for an user when the number of genuine samples is above a defined threshold \cite{eskander:13}.

Yilmaz \cite{yilmaz:16} also propose a hybrid approach that combines WI and WD results. Using the main ideas from the WD and WI approaches, aiming to learn the importance of different dissimilarities, the writer-independent classifier is trained with dissimilarity vectors of query and reference signatures of all users. In its turn, the writer-dependent classifiers are trained separately for each user, to learn to differentiate genuine signatures and forgeries. The results are then combined using a score-level fusion of these complementary classifiers with different local features \cite{yilmaz:16}.

\section{Proposed Method}
\label{sec:proposed_method}

In the context of offline Handwritten Signature Verification, Hafemann et al. \cite{hafemann:17} have achieved significant improvements in results by using a deep CNN to extract writer-independent representative features from signatures (specifically, the 2048 features were obtained from the FC7 layer of the deep CNN). In  Hafemann et al. proposal, the deep CNN features are  used to train a writer-dependent SVM classifier for each user of the system. In contrast, in this paper a writer-independent (WI) SVM classifier is proposed to perform the signature verification.

WI systems use the dissimilarity between each questioned signature and the reference signatures to perform the authentication. To this end, the proposed WI approach employs the dichotomy transformation, which allows to transform $K$-class pattern recognition problems, where $K$ is a large or unspecified value, into a 2-class problem (in this case, classify a handwriting sample into genuine or forgery) \cite{rivard:13}. In this context, the single SVM classifier is trained for all users from this dichotomy space.

Among the advantages of using a WI model are the following: (i) it allows to exploit a system with only one signature per user, (ii) in contrast to WD systems, the number of users in WI is of little consequence, as input feature vectors are transformed into a distance space between signatures, (iii) the writers populating used during the verification do not necessarily need to be enrolled to the system \cite{rivard:13}.

\subsection{Dichotomy Transformation}

The Dichotomy Transformation (DT), proposed by Cha and Srihari \cite{cha:00}, transforms a multi-class problem into a binary problem (two classes). 
To this end, the vector of distances of every feature by the same writer samples are computed and categorized as a \textit{within author} distance (which is denoted by $x_+$). On the other hand, the distance vectors from samples of different writers are categorized as \textit{between author} distance (which is denoted by $x_-$). Let $d_{ij}$ denote the $j'th$ sample of the $i'th$ writer \cite{cha:00}.


\vspace{-0.5cm}
\begin{multline}
x_+ = u(d_{ij} - d_{ik}) \; \\ 
\shoveleft where \; i = 1 \; to \; n; j,k = 1 \; to \; m \; and \; j \neq k
\end{multline}
\vspace{-0.5cm}
\begin{multline}
x_- = u(d_{ij} - d_{kl}) \; \\
\shoveleft where \; i, k = 1 \; to \; n \;  and \; j \neq k; j,l = 1 \; to \; m
\end{multline}

\noindent where $n$ is the number of writers, $m$ is the number of samples per writer, $u$ is the absolute value in the distance domain resulting from the dichotomy transformation. 

The following properties are desired  with the dichotomy transformation: (i) all distances between samples from the same writer in the feature domain should belong to the \textit{within class} in the transformed space (that is, stay close to the origin in the dissimilarity representation space), and (ii) all distances between two different classes in the feature domain should belong to the \textit{between class} in the transformed space (that is, stay away from the origin in the dissimilarity representation space) \cite{cha:00}. 

This is not always the case. One disadvantage of the dichotomy transformation is that perfectly clustered writers in the feature domain may not be perfectly dichotomized in the distance domain \cite{cha:00}. In other words, the broader the spread of the feature distributions among the writers, the less the dichotomizer is able to detect real differences between similar signatures \cite{rivard:13}.

Other properties that are worth highlighting are as follows: (i) the dichotomy transformation affects the geometry of data distributions. So, if multiple boundaries are needed to separate the classes in the feature space, only one is needed in the distance space; (ii) the vectors resulting from the DT are always nonnegative since they consist of distances transformed in absolute values; and (iii) since each sample in the dichotomy space is formed by the distance of each pair of signatures, the limited number of samples is no longer a problem \cite{rivard:13}.

\section{Experiments}
\label{sec:experiments}

\subsection{Datasets}

The experiments are carried out using GPDS and Brazilian PUC-PR datasets. Table \ref{tab:summary} summarizes these datasets.

\newcolumntype{C}[1]{>{\centering\arraybackslash}p{#1}}

\begin{table}[!htb]
\caption{Summary of the used datasets.}
\label{tab:summary}
\scriptsize
\centering

\begin{tabular}{ccC{1cm}c}
\hline
Dataset Name & Users & Genuine signatures (per user) & Forgeries per user \\ 
\hline
Brazilian (PUC-PR) & 60 + 108 & 40  &  10 simple, 10
skilled \\ 
GPDS Signature 960 & 881 & 24  &  30 \\ 

\hline

\end{tabular}
\end{table}

In the first set of experiments the Brazilian PUC-PR dataset is considered. The data segmentation was done as in the paper by Rivard et al. \cite{rivard:13}. The development set $D$ is formed by users 61 to 168 and the exploitation dataset $\varepsilon$ by users 1 to 60.
So, to generate the \textit{within class} of the learning set $L$, 30 randomly selected genuine signatures for each writer from the development dataset $D$ are used, resulting in $108 \cdot \frac{30 \cdot 29}{2} = 46,980$ distance vectors. Aiming to generate a balanced dataset, the \textit{between class} counterexamples of the learning set $L$ are obtained as follows: for each writer, the dichotomy transformation is applied to 29 genuine signatures (references signatures) against 15 random forgeries, each one selected from a genuine signature of 15 different writers, giving $ 108 \cdot 29 \cdot 15 = 46,980$ distance vectors \cite{rivard:13}. 
To perform the verification, the reference set $R$ is composed of 30 randomly selected genuine signatures from each writer of the exploitation dataset $\varepsilon$ (writers from this dataset are unknown to the verification system). The questioned set $Q$ is composed of the 10 remaining genuine signatures, 10 simple and 10 skilled forgeries from each writer plus 10 random forgeries, each one selected from a genuine signature of 10 different writers \cite{rivard:13}.

The metrics used to evaluate the performance are based on Table 10 from \cite{hafemann:17}. 
The following metrics are used: (i) False Rejection Rate ($FRR$), which represents the percentage of genuine signatures that are rejected by the system, (ii) False Acceptance Rate ($FAR$), represents the percentage of forgeries that are accepted (it can be computed for each random, simple and skilled forgeries), (iii) Average Error Rate ($AER$), is the average error considering $FRR$, $FAR_{random}$, $FAR_{simple}$, $FAR_{skilled}$.  (iv) Equal Error Rate ($EER$), is the error obtained when $FRR$ = $FAR$ \cite{hafemann_review:17}.

The comparative analysis for the GPDS dataset are based on values from table 7 of Hafemann et al. paper \cite{hafemann:17}. 
For the GPDS-160 development set $D$ is formed by users 161 to 881 and the exploitation dataset $\varepsilon$ by user 1 to 160.
In turns, for the GPS-300 development set $D$ is formed by users 301 to 881 and the exploitation dataset $\varepsilon$ by user 1 to 300.
In this dataset, the \textit{within class} of the learning set $L$ is generated from 12 randomly selected genuine signatures for each writer from the development dataset $D$, which results in $721 \cdot \frac{12 \cdot 11}{2} = 47,586$ and $581 \cdot \frac{12 \cdot 11}{2} = 38,346$ distance vectors, respectively, for the GPDS-160 and the GPDS-300 datasets.
The \textit{between class} is formed by using 11 genuine signatures (references signatures) against 6 random forgeries, each one selected from a genuine signature of 6 different writers, resulting in $721 \cdot 11 \cdot 6 = 47,586$ and $581 \cdot 11 \cdot 6 = 38,346$ distance vectors, respectively, for the GPDS-160 and the GPDS-300 datasets. To perform the verification, the reference set $R$ is composed of 12 randomly selected genuine signatures from each writer of the exploitation database $\varepsilon$. The questioned set $Q$ is composed by 10 of the remaining genuine signatures and 10 skilled forgeries from each writer plus 10 random forgeries, each one selected from a genuine signature of 10 different writers (trying to follow the segmentation from Rivard et al. paper \cite{rivard:13}). 

In both datasets, the writer-independent SVM classifier is trained using the data from the learning set $L$ and the verification is performed using the questioned $Q$ and the reference $R$ sets.

\subsection{Experimental setup}

Support Vector Machines (SVM) have been widely used for both WD and WI signature verification tasks. This indicates that it is one of the most effective classifiers for this problem \cite{hafemann_review:17}.

In this paper, the SVM is used as writer-independent classifier with the following settings: $RBF$ kernel, $\gamma = 2^{-11}$ and $C=1.0$ (the training data is balanced, so there is no need to use different weights for the positive and negative class) \cite{hafemann:17}. 

All data were randomly selected and a different SVM was trained for each replication (five replications for each experimental configuration were performed).

The point of intersection between $FRR$ and $FAR_{skilled}$ curves is used as a global threshold. The $EER$ metric is also computed based on an user threshold, as presented in Hafemann et al. \cite{hafemann:17}.



\subsection{Results and discussion}

The results are organized as follows: (i) initially, the number of signatures in the reference set $R$ is fixed and an analysis of which partial decisions fusion rule is the best (functions max, mean, median and min were tested). (ii) In the sequence, the analysis about the influence of the number of signatures used in the reference set $R$. (iii) Finally, the comparison with the state-of-the-art for the used datasets (as the present work uses SigNet features, Hafemann et al. results are also presented using only these features \cite{hafemann:17}).

\subsubsection{Brazilian PUC-PR dataset}

\paragraph{Partial decisions fusion rule analysis}
To measure the impact of the partial decisions fusion rule, in this section  the number of references per user is fixed in 30 and different functions are tested. The functions used in the combinations were: (i) Mean, (ii) Max, (iii) Median and (iv) Min.


\begin{figure}[!htb]
\centering
  \includegraphics[width=\columnwidth]{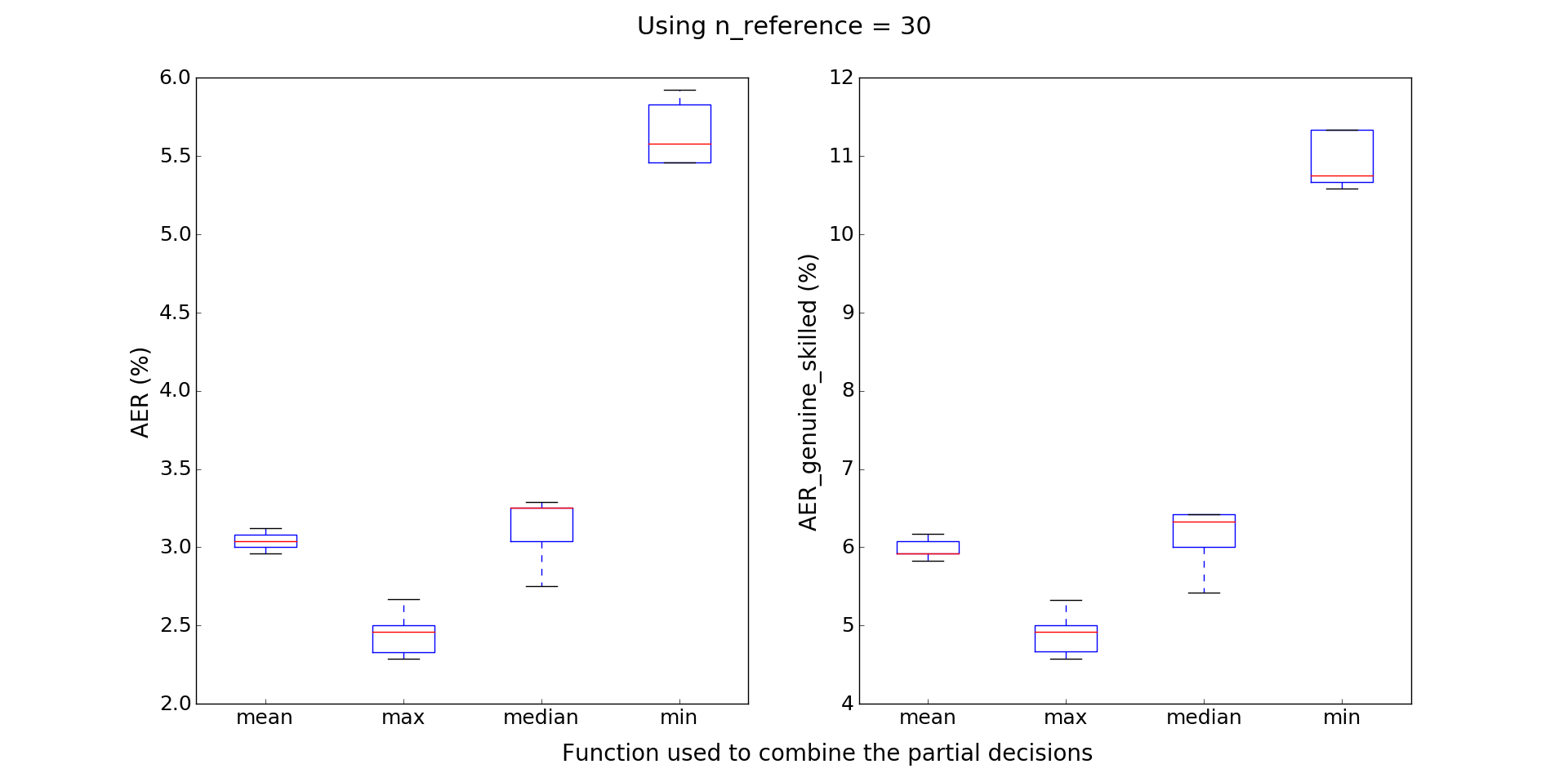}
  \caption{Boxplots for $AER$ (left) and $AER_{genuine+skilled}$ (right) metrics on the Brazilian PUC-PR dataset, using n\_reference = 30.}
  \label{fig:varying_decisionFusion}
\end{figure}

Figure \ref{fig:varying_decisionFusion} presents the boxplots for $AER$ and $AER_{genuine+skilled}$ for the tested functions (Max, Mean, Median, Min). As can be seen, the Max function obtained the best results for both metrics. In the opposite direction, the Min function had the worst results in both cases.

The Wilcoxon paired signed-rank test with 5\% significance level for both $AER$ and $AER_{genuine+skilled}$ metrics shows that the max function outperforms the other functions with statistical relevance. 

For the $EER$ metric, the Wilcoxon paired signed-rank test also showed that the Max function is statistically better when compared to the Median and the Min. However, there is no statistical difference to the Mean function.

In Bertolini et al. \cite{bertolini:10} the Max rule also achieved the best results for the considered dataset.

\paragraph{Analysis of the influence of the number of reference signatures}
In the previous section it was shown that the Max function obtained better results when compared to other functions. 
In this section, to measure the impact of the reference set cardinality, the Max function was fixed and the number of references per user was varied. To this end, reference subsets containing [1, 5, 10, 15, 20, 25, 30] randomly selected signatures are used as references for the authentication.

\begin{figure}[!htb]
\centering
  \includegraphics[width=\columnwidth]{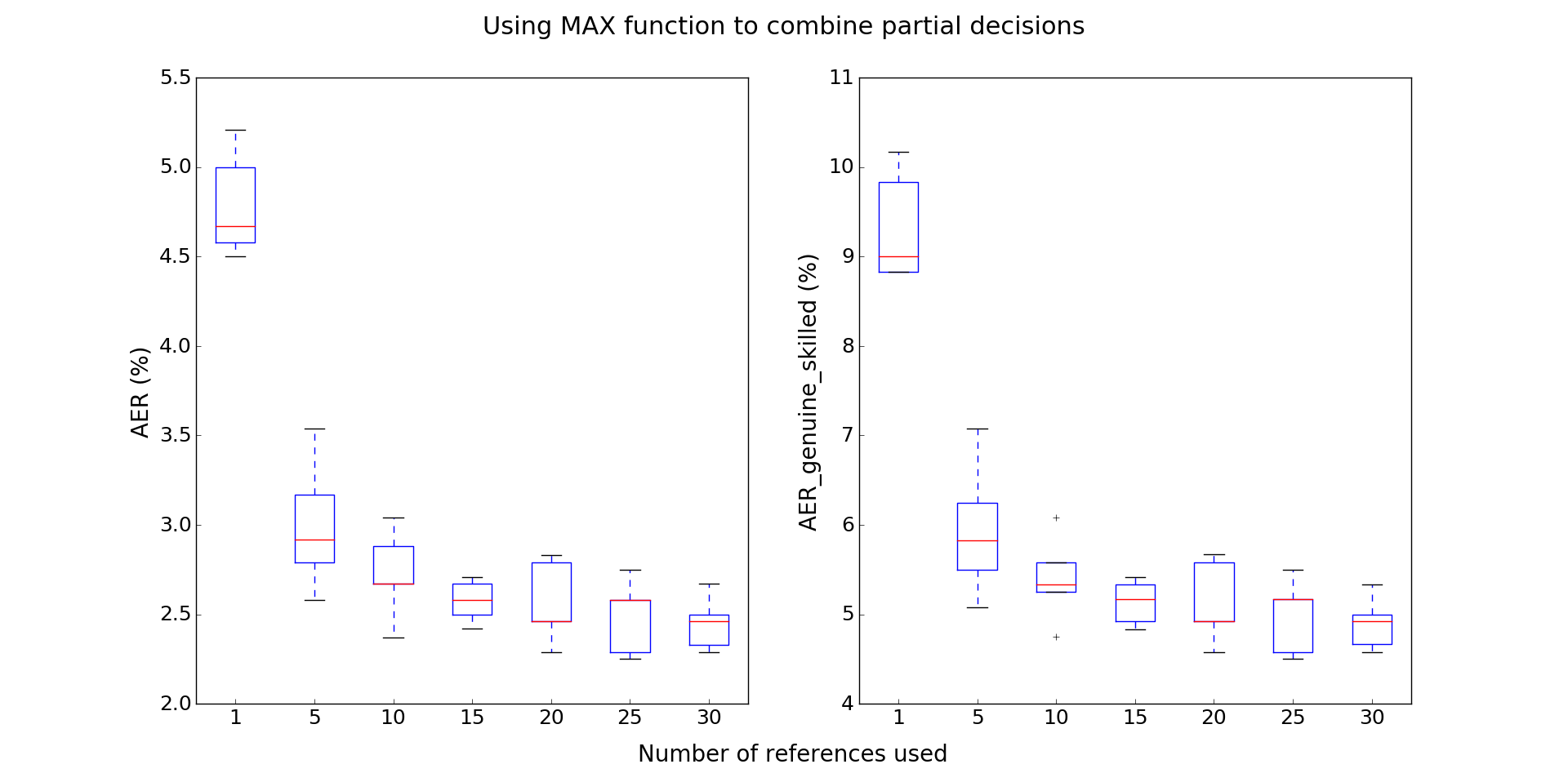}
  \caption{Boxplots for AER (left) and $AER_{genuine+skilled}$ (right) metrics on the Brazilian PUC-PR dataset, using Max function.}
  \label{fig:varying_nReference}
\end{figure}

Figure \ref{fig:varying_nReference} depicts the boxplots for $AER$ and $AER_{genuine+skilled}$ for different reference sizes. As can be observed, using more references per user produces better results (the worst cases are with number of references = 1 and = 5). However, the variation among results decreases as the number of references increases. For instance, the experiments using 15 or 20 references present similar results; the same can be observed with 25 and 30.
  
The Wilcoxon paired signed-rank test for both metrics, using n\_reference = 30 as baseline, shows that results are statistically better only when compared with the cases where number of references = 1 and = 5. There is no statistical difference to n = 10, 15, 20 or 25.

So far, the best results are obtained using the Max function with the highest value for the number of references (in this case, n\_reference = 30) for both global and user threshold scenarios (this settings will be referenced as ``best proposed approach'').


\paragraph{Comparison with the state-of-the-art}
Table \ref{tab:comparison_metrics} presents results obtained by the best proposed approach together with results from table 10 of Hafemann et al. work \cite{hafemann:17}. 

\begin{table*}[!htb]
\caption{Comparison with the state-of-the-art on the Brazilian PUC-PR dataset, using Max function (errors in \%).}
\label{tab:comparison_metrics}
\scriptsize
\centering

\begin{tabular}{cccccccc}
\hline
Reference & \#samples & $FRR$  &  $FAR_{random}$ & $FAR_{simple}$  & $FAR_{skilled}$  &  $AER$  &  $AER_{genuine+skilled}$ \\ 
\hline
Bertolini et al. \cite{bertolini:10} & 15 &  10.16  &  3.16 & 2.8  & 6.48  &  5.65 &  8.32 \\ 
Batista et al. \cite{batista:12} & 30 &  7.5  &  0.33 & 0.5  & 13.5  &  5.46 &  10.5 \\ 
Rivard et al. \cite{rivard:13} & 15  &  11  &  0.0 & 0.19  & 11.15  &  5.59 &  11.08 \\ 
Eskander et al. \cite{eskander:13} & 30 &  7.83  &  0.02 & 0.17  & 13.5  &  5.38 &  10.67 \\ 
Hafemann, Sabourin and Oliveira \cite{hafemann:16} & 15  &  2.17  &  0.17 & 0.50 & 13.00 &  3.96 & 7.59 \\ 
Hafemann et al. \cite{hafemann:17} & 5  &  4.63 (0.55)  &  0.00 (0.00) & 0.35 (0.20)  & 7.17 (0.51)  &  3.04 (0.17) &  5.90 (0.32) \\ 
Hafemann et al. \cite{hafemann:17} & 15  &  1.22 (0.63)  &  0.02 (0.05) & 0.43 (0.09) & 10.70 (0.39)  &  3.09 (0.20) &  5.96 (0.40) \\ 
Hafemann et al. \cite{hafemann:17} & 30  &  0.23 (0.18)  &  0.02 (0.05) & 0.67 (0.08)  & 12.62 (0.22)  &  3.38 (0.06) &  6.42 (0.13) \\ 
Present work (global\_threshold)  & 5 & 5.95 (0.68) & 0.00 (0.00)  & 0.10 (0.08)  & 5.95 (0.68)  &  3.00 (0.33) &  5.95 (0.68) \\ 
Present work (global\_threshold)  & 15 & 5.13 (0.23) & 0.00 (0.00)  & 0.03 (0.07)  & 5.13 (0.23)  & 2.58 (0.11)  & 5.13 (0.23)  \\ 
\textbf{Present work (global\_threshold)}  & 30 & 4.90 (0.27)  &  0.00 (0.00) & 0.00 (0.00)  & 4.90 (0.27)  &  \textbf{2.45 (0.13)} &  \textbf{4.90 (0.27)} \\ 
\hline

\end{tabular}
\end{table*}

As shown in Table \ref{tab:comparison_metrics}, when compared to Hafemann et al. work \cite{hafemann:17}, the results of the best proposed approach are worse in $FRR$ metric, but are better in  $FAR_{skilled}$ metric, in a global scenario. In summary, the results of the present work are better than Hafemann et al. work \cite{hafemann:17} (this can be seen through $AER$ and $AER_{genuine+skilled}$). 
When compared to the other works, the presented approach produces a considerable improvement for all metrics. Keep in mind that the proposed method uses a simpler architecture (using only a single SVM to perform the verification) when compared to others, such as the ones proposed by Bertolini et al. \cite{bertolini:10} or Rivard et al. \cite{rivard:13}, that use ensembles of classifiers.

\begin{table}[!htb]
\caption{Comparison of $EER$ with the state-of-the-art on the Brazilian PUC-PR dataset, using Max function (errors in \%).}
\label{tab:comparison_ERR_userthreshold}
\scriptsize
\centering

\begin{tabular}{cccc}
\hline
Type & Reference & \#samples & $EER$ \\ 
\hline
WD & Hafemann, Sabourin and Oliveira \cite{hafemann:16} & 15  &  4.17 \\ 
WD & Hafemann et al. \cite{hafemann:17} & 5  &  2.92 (0.44) \\ 
WD & Hafemann et al. \cite{hafemann:17} & 15  &  2.07 (0.63) \\ 
WD & Hafemann et al. \cite{hafemann:17} & 30  &  2.01 (0.43) \\ 
WI & Present work (using a global threshold) & 5 & 5.95 (0.68) \\
WI & Present work (using a global threshold) & 15 & 5.13 (0.23) \\
WI & Present work (using a global threshold) & 30 & 4.90 (0.27) \\ 
WI & Present work (using an user threshold)  & 5 & 2.58 (0.72) \\ 
WI & Present work (using an user threshold)  & 15 & 1.70 (0.40) \\ 
WI & \textbf{Present work (using an user threshold)}  & 30 & \textbf{1.48 (0.44)} \\ 
\hline

\end{tabular}
\end{table}

As can be observed in Table \ref{tab:comparison_ERR_userthreshold}, the results of the best proposed approach of this paper obtained slightly superior results in comparison with Hafemann, Sabourin and Oliveira \cite{hafemann:16} and Hafemann et al. \cite{hafemann:17} for the $EER$ metric. It worth notice that the proposed method performs writer-independent authentication and both Hafemann's models operate in a writer-dependent way and, even so, the WI approach was able to improve the results.


\subsubsection{GPDS dataset}
For the GPDS datasets, the same functions used as partial decisions fusion rule were tested. As the highest value for the number of references is 12, reference subsets containing [1, 2, 3, 4, 5, 10, 12] randomly selected signatures were tested as references for the authentication.

\begin{figure}[!htb]
\centering
  \includegraphics[width=\columnwidth]{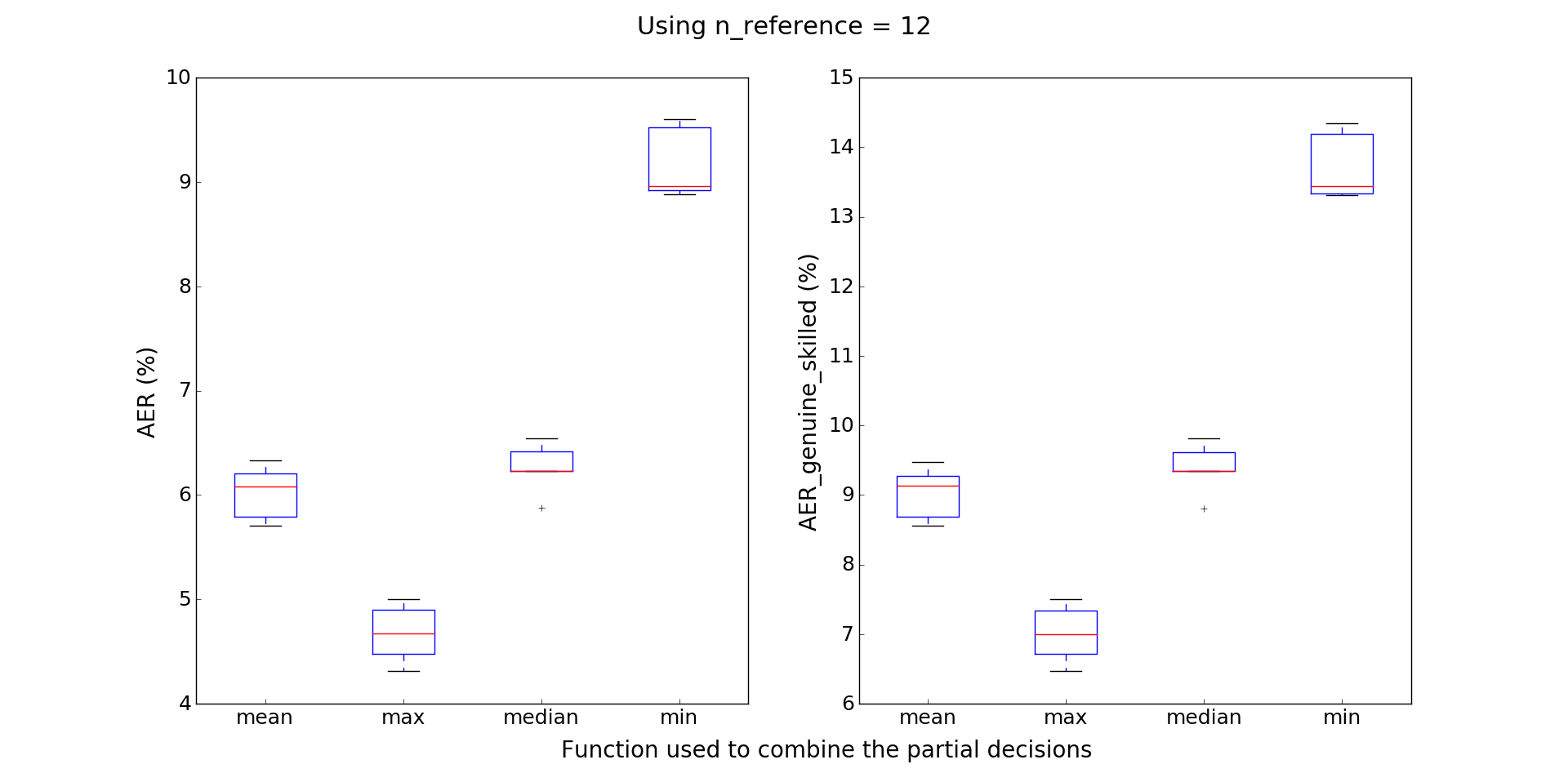}
  \caption{Boxplots for $AER$ (left) and $AER_{genuine+skilled}$ (right) metrics on the GPDS-160 dataset, using n\_reference = 12.}
  \label{fig:gpds160_varying_decisionFusion}
\end{figure}

\begin{figure}[!htb]
\centering
  \includegraphics[width=\columnwidth]{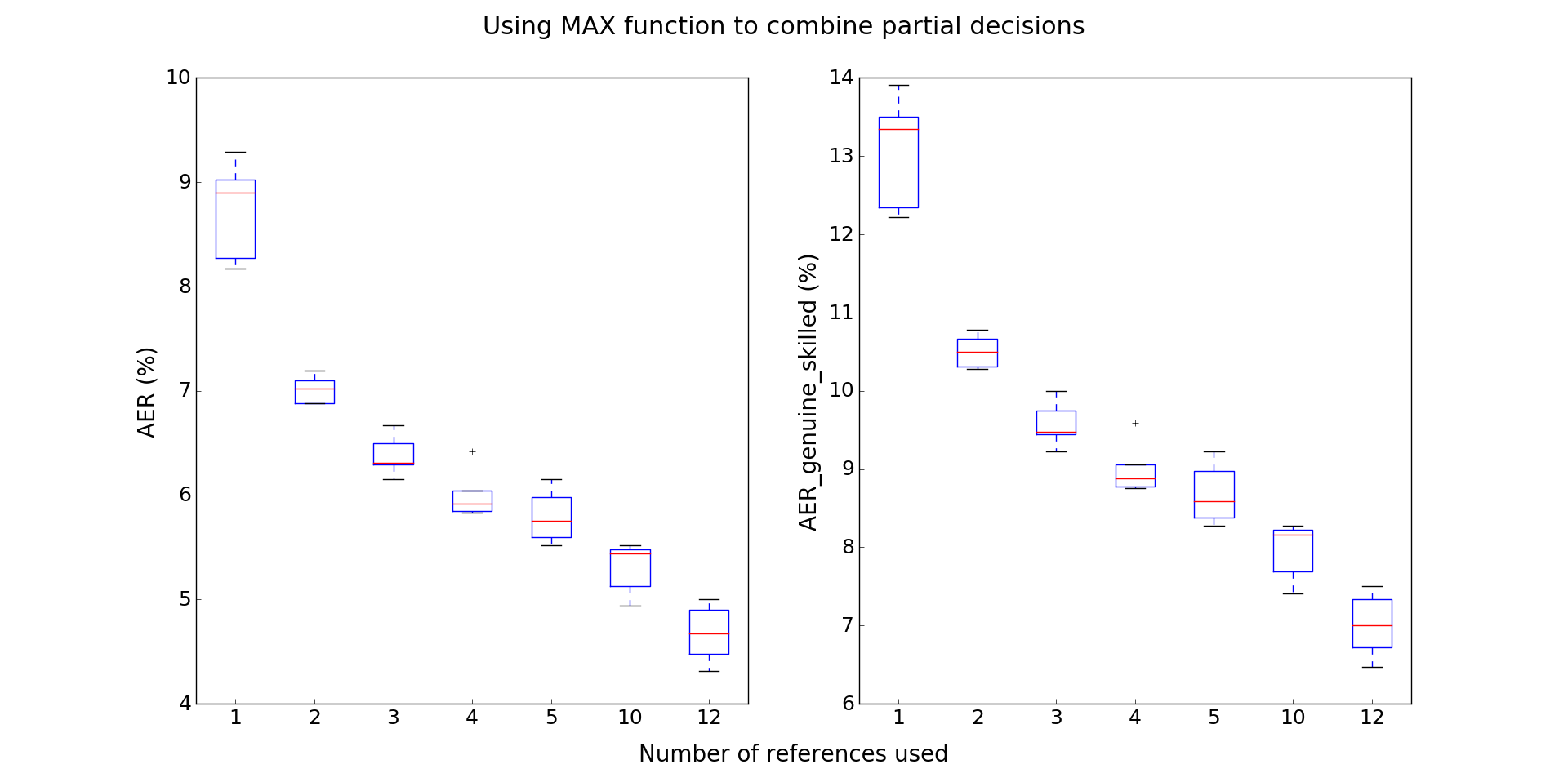}
  \caption{Boxplots for AER (left) and $AER_{genuine+skilled}$ (right) metrics on the GPDS-160 dataset, using Max function.}
  \label{fig:gpds160_varying_nReference}
\end{figure}

As in Brazilian PUC-PR dataset, in the GPDS dataset the best results are obtained using the Max function with the highest value for the number of references (in this case, n\_reference = 12) for both global and user threshold scenarios.  
Figures \ref{fig:gpds160_varying_decisionFusion} and \ref{fig:gpds160_varying_nReference} depict, respectively, the boxplots for $AER$ and $AER_{genuine+skilled}$ metrics (i) when varying the fusion rule and (ii) when we varied the reference sizes, for the GPDS-160 dataset. Similar behavior can be observed for both GPDS-160 and GPDS-300.

For the GPDS-160 dataset, performing the Wilcoxon paired signed-rank test for both metrics: (i) Max function outperforms
the other functions with statistical relevance. (ii) Using n\_reference = 12 is statistically better when compared to the other cases.

Tables \ref{tab:comparison_GPDS_160} and \ref{tab:comparison_GPDS_300} presents the results when the best proposed approach is used and compares the obtained results with those from Table 7 of Hafemann et al. paper \cite{hafemann:17}.

\begin{table}[!htb]
\caption{Comparison of $EER$ with the state-of-the-art on the GPDS-160 dataset, using Max function (errors in \%).}
\label{tab:comparison_GPDS_160}
\scriptsize
\centering

\begin{tabular}{cccc}
\hline
Type & Reference & \#samples & $EER$ \\ 
\hline
WD & Guerbai et al. \cite{guerbai:15} & 12  &  15.07 \\ 
WD & Hafemann, Sabourin and Oliveira \cite{hafemann:16} & 12  & 10.70 \\ 
WI + WD & Yilmaz \cite{yilmaz:16} & 5  &  7.98 \\ 
WI + WD & Yilmaz \cite{yilmaz:16} & 12  &  6.97 \\
WD & Hafemann et al. \cite{hafemann:17} & 5 &  3.23 (0.36) \\ 
WD & \textbf{Hafemann et al. \cite{hafemann:17}} & 12  &  \textbf{2.63 (0.36)} \\ 
WI & Present work (using a global threshold) & 5 & 8.69 (0.36) \\ 
WI & Present work (using a global threshold) & 12  & 7.01 (0.38) \\
WI & Present work (using an user threshold) & 5 & 4.01 (0.39) \\ 
WI & Present work (using an user threshold) & 12  & 2.86 (0.24) \\ 
\hline

\end{tabular}
\end{table}

\begin{table}[!htb]
\caption{Comparison of $EER$ with the state-of-the-art on the GPDS-300 dataset, using Max function (errors in \%).}
\label{tab:comparison_GPDS_300}
\scriptsize
\centering

\begin{tabular}{cccc}
\hline
Type & Reference & \#samples & $EER$ \\ 
\hline
WD & Soleimani et al. \cite{soleimani:16} & 10  &  20.94 \\ 
WD & Hafemann, Sabourin and Oliveira \cite{hafemann:16} & 12  & 12.83 \\ 
WD & Hafemann et al. \cite{hafemann:17} & 5  &  3.92 (0.18) \\ 
WD & \textbf{Hafemann et al. \cite{hafemann:17}} & 12  &  \textbf{3.15 (0.18)} \\ 
WI & Present work (using a global threshold) & 5 & 9.05 (0.34) \\ 
WI & Present work (using a global threshold) & 12 & 7.96 (0.26) \\ 
WI & Present work (using an user threshold) & 5 & 4.40 (0.34) \\ 
WI & Present work (using an user threshold) & 12 & 3.34 (0.22) \\ 
\hline

\end{tabular}
\end{table}

As can be observed in Table \ref{tab:comparison_GPDS_160}, for the GPDS-160, our best proposed approach was able to outperform Guerbai et al. \cite{guerbai:15} and Hafemann, Sabourin and Oliveira \cite{hafemann:16} using both global and user thresholds. Also, it was able to outperform Yilmaz \cite{yilmaz:16} (with 5 and 12 samples) using user thresholds. 
As in Brazilian PUC-PR dataset, in GPDS-160 dataset, the proposed approach was able to achieve better results with a simpler architecture. While we are using only a single SVM to perform the verification, Yilmaz \cite{yilmaz:16} uses an ensemble of WI and WD classifiers.

For the GPDS-300, as presented in Table \ref{tab:comparison_GPDS_300}, our best proposed approach was able to outperform Soleimani et al. \cite{soleimani:16} and Hafemann, Sabourin and Oliveira \cite{hafemann:16} using both global and user thresholds. 

However, for both datasets, the best proposed approach using an user threshold obtained slightly inferior results in comparison with Hafemann et al. WD model \cite{hafemann:17} for the $EER$ metric. 


\section{Conclusion}
\label{sec:conclusion}
This paper has introduced an approach for writer-independent offline signature verification that uses the dissimilarity representation of the deep CNN features from \cite{hafemann:17} and a single SVM as a writer-independent classifier to authenticate handwritten signatures. 

The experiments showed that, in general, for the tested datasets, the best results are obtained using the Max function as the partial decisions fusion rule with the highest value for the number of references for both global and user threshold scenarios.

Moreover, in the global threshold scenario, the proposed approach was able to outperform Hafemann et al. \cite{hafemann:17} in the Brazilian dataset. For the user threshold scenario, the proposed approach was able to obtain performance comparable to Hafemann et al. \cite{hafemann:17}. This was so even with the proposed method performing writer-independent authentication and Hafemann et al. \cite{hafemann:17} operating in a writer-dependent way. In the Brazilian dataset the proposed work was slightly superior and in the GPDS dataset slightly inferior. However, for both datasets, the proposed approach was able to outperform other methods from the literature that use WD classification and the WI dissimilarity representation with different features and more complex classification architectures (for instance, ensembles of classifiers). 

Future research will include the study of feature and prototype selection in the dissimilarity space, adaptation of WI classifier over time and the writer-dependent decision threshold. 

 
\bibliographystyle{IEEEtran}
\bibliography{IEEEabrv,mybibliography}

\begin{thebibliography}{10}
\providecommand{\url}[1]{#1}
\csname url@samestyle\endcsname
\providecommand{\newblock}{\relax}
\providecommand{\bibinfo}[2]{#2}
\providecommand{\BIBentrySTDinterwordspacing}{\spaceskip=0pt\relax}
\providecommand{\BIBentryALTinterwordstretchfactor}{4}
\providecommand{\BIBentryALTinterwordspacing}{\spaceskip=\fontdimen2\font plus
\BIBentryALTinterwordstretchfactor\fontdimen3\font minus
  \fontdimen4\font\relax}
\providecommand{\BIBforeignlanguage}[2]{{%
\expandafter\ifx\csname l@#1\endcsname\relax
\typeout{** WARNING: IEEEtran.bst: No hyphenation pattern has been}%
\typeout{** loaded for the language `#1'. Using the pattern for}%
\typeout{** the default language instead.}%
\else
\language=\csname l@#1\endcsname
\fi
#2}}
\providecommand{\BIBdecl}{\relax}
\BIBdecl

\bibitem{hafemann_review:17}
L.~G. Hafemann, R.~Sabourin, and L.~S. Oliveira, ``Offline handwritten
  signature verification—literature review,'' in \emph{Image Processing
  Theory, Tools and Applications (IPTA), 2017 Seventh International Conference
  on}.\hskip 1em plus 0.5em minus 0.4em\relax IEEE, 2017, pp. 1--8.

\bibitem{kumar:16}
A.~Kumar and K.~Bhatia, ``A survey on offline handwritten signature
  verification system using writer dependent and independent approaches,'' in
  \emph{Advances in Computing, Communication, \& Automation (ICACCA)(Fall),
  International Conference on}.\hskip 1em plus 0.5em minus 0.4em\relax IEEE,
  2016, pp. 1--6.

\bibitem{hafemann:17}
L.~G. Hafemann, R.~Sabourin, and L.~S. Oliveira, ``Learning features for
  offline handwritten signature verification using deep convolutional neural
  networks,'' \emph{Pattern Recognition}, vol.~70, pp. 163--176, 2017.

\bibitem{rivard:13}
D.~Rivard, E.~Granger, and R.~Sabourin, ``Multi-feature extraction and
  selection in writer-independent off-line signature verification,''
  \emph{International Journal on Document Analysis and Recognition (IJDAR)},
  vol.~16, no.~1, pp. 83--103, 2013.

\bibitem{batista:12}
L.~Batista, E.~Granger, and R.~Sabourin, ``Dynamic selection of
  generative--discriminative ensembles for off-line signature verification,''
  \emph{Pattern Recognition}, vol.~45, no.~4, pp. 1326--1340, 2012.

\bibitem{guerbai:15}
Y.~Guerbai, Y.~Chibani, and B.~Hadjadji, ``The effective use of the one-class
  svm classifier for handwritten signature verification based on
  writer-independent parameters,'' \emph{Pattern Recognition}, vol.~48, no.~1,
  pp. 103--113, 2015.

\bibitem{bertolini:10}
D.~Bertolini, L.~S. Oliveira, E.~Justino, and R.~Sabourin, ``Reducing forgeries
  in writer-independent off-line signature verification through ensemble of
  classifiers,'' \emph{Pattern Recognition}, vol.~43, no.~1, pp. 387--396,
  2010.

\bibitem{eskander:13}
G.~S. Eskander, R.~Sabourin, and E.~Granger, ``Hybrid
  writer-independent--writer-dependent offline signature verification system,''
  \emph{IET biometrics}, vol.~2, no.~4, pp. 169--181, 2013.

\bibitem{yilmaz:16}
M.~B. Y{\i}lmaz and B.~Yan{\i}ko{\u{g}}lu, ``Score level fusion of classifiers
  in off-line signature verification,'' \emph{Information Fusion}, vol.~32, pp.
  109--119, 2016.

\bibitem{cha:00}
S.-H. Cha and S.~N. Srihari, ``Writer identification: statistical analysis and
  dichotomizer,'' in \emph{Joint IAPR International Workshops on Statistical
  Techniques in Pattern Recognition (SPR) and Structural and Syntactic Pattern
  Recognition (SSPR)}.\hskip 1em plus 0.5em minus 0.4em\relax Springer, 2000,
  pp. 123--132.

\bibitem{hafemann:16}
L.~G. Hafemann, R.~Sabourin, and L.~S. Oliveira, ``Writer-independent feature
  learning for offline signature verification using deep convolutional neural
  networks,'' in \emph{Neural Networks (IJCNN), 2016 International Joint
  Conference on}.\hskip 1em plus 0.5em minus 0.4em\relax IEEE, 2016, pp.
  2576--2583.

\bibitem{soleimani:16}
A.~Soleimani, B.~N. Araabi, and K.~Fouladi, ``Deep multitask metric learning
  for offline signature verification,'' \emph{Pattern Recognition Letters},
  vol.~80, pp. 84--90, 2016.

\end{thebibliography}

\end{document}